\documentclass[runningheads]{llncs}
\usepackage{graphicx}

\usepackage{tikz}
\usepackage{comment}
\usepackage{amsmath,amssymb} 
\usepackage{color}

\newcommand{\ie}{\textit{i.e.}}
\newcommand{\eg}{\textit{e.g.}}
\newcommand{\etal}{\textit{et al. }}

\newcommand{\norm}[1]{\left\lVert#1\right\rVert}
\newcommand{\inlineimg}[1]{\raisebox{-0.2\baselineskip}{\includegraphics[height=0.95\baselineskip]{#1.png}}}

\newcommand{\algname}{\textsc{EqInv }}
\newcommand{\algnamens}{\textsc{EqInv}}

\usepackage{booktabs}
\usepackage{multirow}
\usepackage{makecell}
\usepackage{graphicx}
\usepackage{diagbox}
\usepackage{tabu}
\usepackage{rotating}
\usepackage{array}
\usepackage{float}
\usepackage{enumitem}
\usepackage{pifont}

\usepackage{mathrsfs}
\usepackage{caption}
\usepackage{wrapfig}
\usepackage{bbm}
\usepackage{colortbl}
\usepackage{bbding}

\usepackage{bm}
\usepackage[pagebackref,breaklinks,colorlinks]{hyperref}

\newcommand\independent{\protect\mathpalette{\protect\independenT}{\perp}}
\def\independenT#1#2{\mathrel{\rlap{$#1#2$}\mkern2mu{#1#2}}}

\definecolor{mygray}{gray}{0.9}

\newlength\savewidth
  
\newcolumntype{x}[1]{>{\centering\arraybackslash}p{#1pt}}
\newcolumntype{y}[1]{>{\raggedright\arraybackslash}p{#1pt}}
\newcolumntype{z}[1]{>{\raggedleft\arraybackslash}p{#1pt}}

\definecolor{ABOVE}{HTML}{FF0000}  %
\definecolor{BELOW}{HTML}{39b54a}  %

\usepackage[accsupp]{axessibility}  

\begin{document}

\pagestyle{headings}
\mainmatter
\def\ECCVSubNumber{3535}  

\title{Equivariance and Invariance Inductive Bias for Learning from Insufficient Data} 

\titlerunning{\textsc{EqInv} for Learning from Insufficient Data}
%
\authorrunning{T. Wang, \emph{et al.}}
\author{Tan Wang$^{1}$ \quad 
Qianru Sun$^{2}$ \quad 
Sugiri Pranata$^{3}$ \quad 
Karlekar Jayashree$^{3}$ \quad 
Hanwang Zhang$^{1}$
}
\institute{\footnotesize $^1$Nanyang Technological University \quad
$^2$Singapore Management University \quad
$^3$Panasonic R\&D Center Singapore \\
\email{\tt\footnotesize {\{tan317,hanwangzhang\}@ntu.edu.sg \quad qianrusun@smu.edu.sg \quad 
\{sugiri.pranata,karlekar.jayashree\}@sg.panasonic.com}
}}

\maketitle

\begin{abstract}
We are interested in learning robust models from insufficient data, without the need for any externally pre-trained checkpoints. First, compared to sufficient data, we show why insufficient data renders the model more easily biased to the limited training environments that are usually different from testing. For example, if all the training \texttt{swan} samples are ``white'', the model may wrongly use the ``white'' environment to represent the intrinsic class \texttt{swan}.  Then, we justify that \textbf{equivariance} inductive bias can retain the class feature while \textbf{invariance} inductive bias can remove the environmental feature, leaving the class feature that generalizes to any environmental changes in testing.
To impose them on learning, for equivariance, we demonstrate that any off-the-shelf contrastive-based self-supervised feature learning method can be deployed; for invariance, we propose a class-wise invariant risk minimization (IRM) that efficiently tackles the challenge of missing environmental annotation in conventional IRM. State-of-the-art experimental results on real-world benchmarks (VIPriors, ImageNet100 and NICO) validate the great potential 
of \textbf{equivariance} and \textbf{invariance}
in data-efficient learning. The code is available at \small{\url{https://github.com/Wangt-CN/EqInv}}.
\end{abstract}

\section{Introduction}
\label{sec:intro}

\begin{figure}[t]
\captionsetup{font=footnotesize,labelfont=footnotesize}
    \centering
    \includegraphics[width=1.0\textwidth]{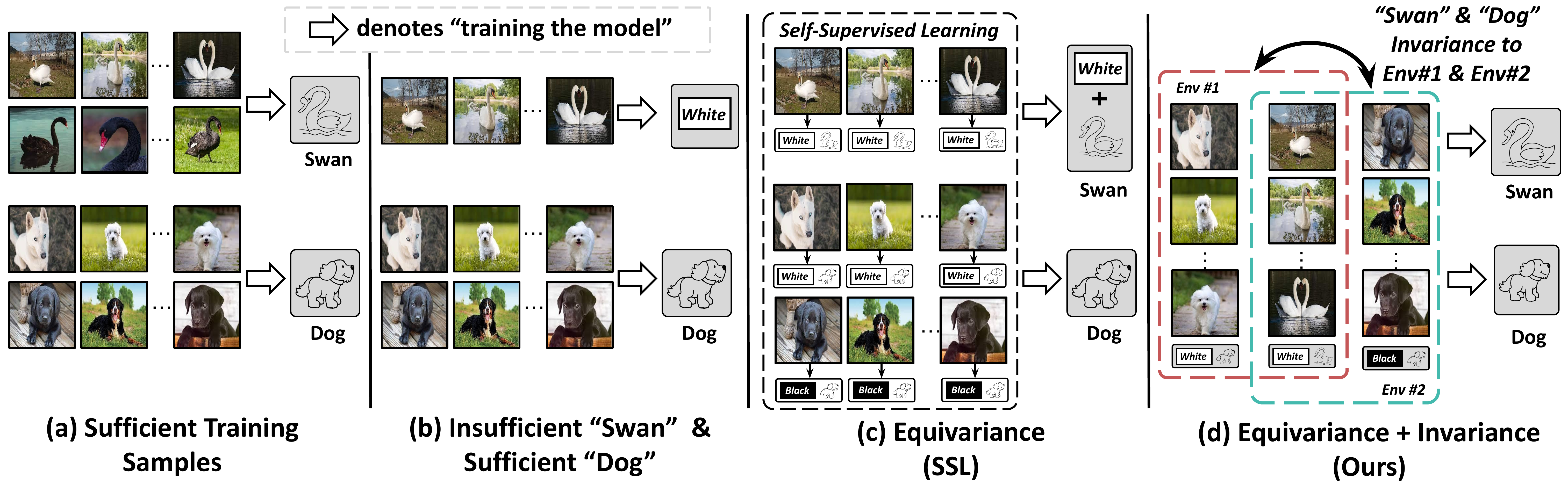}
    \vspace{-5mm}
    \caption{Illustration of how the proposed equivariance and invariance inductive biases help learning from insufficient data. Cartoon figures such as \inlineimg{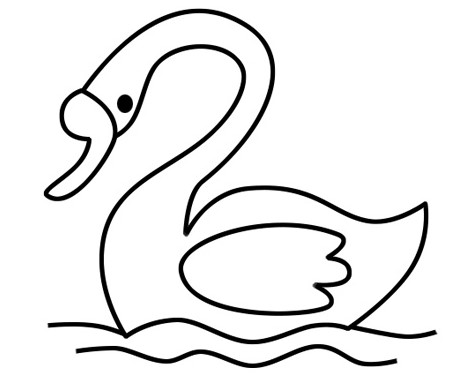} denote the class feature. Boxed words such as \inlineimg{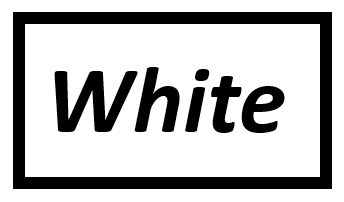} denote environmental features. Grey-boxed figures denote the learned model. For simple illustration, we omit the environment as background.}
    \label{fig:1}
\end{figure}

Data is never too big.  As illustrated in Fig.~\ref{fig:1}~(a), if we have sufficiently large training sample size of \texttt{swan} and \texttt{dog}, \eg, dogs and cats in any environment such as different colors, shapes, poses, and backgrounds, by using a conventional softmax cross-entropy based  ``\texttt{swan} vs. \texttt{dog}''  classifier, we can obtain a ``perfect'' model that discards the \emph{shared} \textbf{environmental} features but retains the \emph{discriminative} \textbf{class} features. 
The underlying common sense is that if the model has seen any ``case'' in training, the testing data is merely a seen IID subset of the training data, yielding testing accuracy as good as training~\cite{vapnik1992principles}.

In this paper, we are interested in learning from insufficient data. Besides the common motivation that collecting data is expensive, we believe that how to narrow the performance gap between insufficient and sufficient data is the key to tackling the non-IID challenge in machine generalization---even if the training data is sufficient, the testing can still be out of the training distribution (OOD)~\cite{he2021towards,recht2019imagenet,shen2021towards}. After all, we can always frame up exceptional testing samples that fail the trained model~\cite{hendrycks2021natural,goodfellow2014explaining}. Note that \textbf{different from few-shot learning} which widely adopts pre-training on large-scale training set~\cite{sung2018learning,snell2017prototypical,sun2019meta}, our task does not allow using any externally pre-trained checkpoint and backbone\footnote[1]{\scriptsize{See https://vipriors.github.io/ for details.}}.

Fig.~\ref{fig:1}~(b) illustrates why insufficient data hurts generalization. Without loss of generality, we conduct a thought experiment that we have limited \texttt{swan} only in ``white'' color environment while sufficient \texttt{dog} in diverse environments. So, we can expect that the ``dog'' feature will still be extracted to represent \texttt{dog} model, but the ``white'' feature will be recklessly learned to represent \texttt{swan}. 
This is because  training \texttt{swan} model by using either ``swan'' or ``white'' feature yields the similar training risk: 1) if the former, the training loss is minimized as in the perfect case of Fig.~\ref{fig:1}~(a); 
2) if the latter, the only training error possible would be misclassifying ``white dog'' as \texttt{swan}. However, it can be easily corrected in practice, \eg, by discriminatively training a sample-to-model distance prior that $\|\mathbf{z}_{\textrm{dog}}\|>\|\mathbf{z}_{\textrm{white}}\|$, where $\mathbf{z}$ denotes the feature vector\footnote[2]{\scriptsize{The distance between the ``white dog'' sample vector $(\mathbf{z}_{\textrm{white}}, \mathbf{z}_{\textrm{dog}})$ and the \texttt{swan} model vector $(\mathbf{z}_{\textrm{white}}, \mathbf{0})$ is: 
$\|(\mathbf{z}_{\textrm{white}}, \mathbf{z}_{\textrm{dog}})-(\mathbf{z}_{\textrm{white}}, \textbf{0})\| = \|(\textbf{0}, \mathbf{z}_{\textrm{dog}})\| = \|\mathbf{z}_{\textrm{dog}}\|$; 
similarly, we have the distance between ``white dog'' and \texttt{dog} model as $\|\mathbf{z}_{\textrm{white}}\|$.}}.

Why, under the same training risk, does the \texttt{swan} model prefer ``white'' but not ``swan'' feature? First, feature extraction in deep network follows a bottom-up, low-level to high-level fashion~\cite{lecun2015deep}---``simple'' features such as color can be easily learned at lower layers, while ``complex'' features such as object parts will be only emerged in higher layers~\cite{selvaraju2017grad,zhou2016learning,zeiler2014visualizing}. Second, the commonly used cross-entropy loss encourages models to stop learning once ``simple'' features suffice to minimize the loss~\cite{geirhos2020shortcut,geirhos2018imagenet}.
As a result, ``swan'' features like ``feather'', ``beak'', and ``wing'' will be lost after training. Such mechanism is also well-known as the shortcut bias~\cite{geirhos2020shortcut} or spurious correlation in causality literature~\cite{pfister2019invariant,wang2021causal}. We will provide formal justifications in Section~\ref{sec:3_1}.

By comparing the difference between Fig.~\ref{fig:1}~(a) and Fig.~\ref{fig:1}~(b), we can see that the crux of improving the generalization of insufficient data is to recover the missing ``swan'' class feature while removing the ``white'' environmental feature.
To this end, we propose two inductive biases to guide the learning: \emph{equivariance} for class preservation and \emph{invariance} for environment removal.

\noindent\textbf{Equivariance}. This prior requires that the feature representation of a sample should be equivariant to its semantic changes, \eg, any change applied to the sample should be faithfully reflected in the feature change (see Appendix for the mathematical definition). Therefore, if we impose a contrastive loss for each sample feature learning, we can encourage that different samples are mapped into different features (see Section~\ref{sec:3_2} for a detailed analysis and our choice). As illustrated in Fig.~\ref{fig:1}~(c), equivariance avoids the degenerated case in Fig~\ref{fig:1}~(b), where all ``white swan'' samples collapse to the same ``white'' feature. Thus, for a testing ``black swan'', the retained ``swan'' feature can win back some \texttt{swan} scores despite losing the similarity between ``black'' and ``white''. It is worth noting that the equivariance prior may theoretically shed light on the recent findings that self-supervised learning features can improve model robustness~\cite{hendrycks2019using,saito2020universal,hendrycks2020pretrained,wen2021toward}. We will leave it as future work.

\noindent\textbf{Invariance}. Although equivariance preserves all the features, due to the limited environments, the \texttt{swan} model may still be confounded by the ``white'' environment, that is, a testing ``black swan'' may still be misclassified as \texttt{dog}, \eg, when $\|(\mathbf{z}_\textrm{black}-\mathbf{z}_\textrm{white},\mathbf{z}_\textrm{swan}-\mathbf{z}_\textrm{swan})\|>\|(\mathbf{z}_\textrm{black}-\mathbf{0}, \mathbf{z}_\textrm{swan}-\mathbf{z}_\textrm{dog})\|$. Inspired by invariant risk minimization~\cite{arjovsky2019invariant} (IRM) that removes the environmental bias by imposing the environmental invariance prior (Section~\ref{sec:3_3}), as shown in Fig.~\ref{fig:1}~(d), if we split the training data into two environments: ``white swan'' vs. ``white dog'' and ``white swan'' vs. ``black dog'', we can learn a common classifier (\ie, a feature selector) that focuses on the ``swan'' and ``dog'' features, which are the \emph{only invariance} across the two kinds of color environments---one is identical as ``white'' and the other contains two colors. 
Yet, conventional IRM requires environment annotation, which is however impractical. To this end, in Section~\ref{sec:4}, we propose \textbf{class-wise IRM} based on contrastive objective that works efficiently without the need for the annotation. We term the overall algorithm of using the two inductive biases, \ie, \textbf{equivariance} and \textbf{invariance}, as $\textsc{EqInv}$.

We validate the effectiveness of $\textsc{EqInv}$ on three real-world visual classification benchmarks: 1) VIPriors ImageNet classification~\cite{bruintjes2021vipriors}, where we evaluate 10/20/50 samples per class; 2) NICO~\cite{he2021towards}, where the training and testing environmental distributions are severely different;  and 3) ImageNet100~\cite{tian2019contrastive} which denotes the case of sufficient training data. On all datasets, we observe significant improvements over baseline learners. Our $\textsc{EqInv}$ achieves a new single-model state-of-the-art on test split: 52.27\% on VIPriors-50 and 64.14\% on NICO.

\section{Related Work}
\noindent\textbf{Visual Inductive Bias}. For a learning problem with many possible solutions, inductive bias is a heuristic prior knowledge that regularizes the learning behavior to find a better solution~\cite{mitchell1980need}. It is ubiquitous in any modern deep learning models: from the shallow MLP~\cite{loshchilov2017decoupled} to the complex deep ResNet~\cite{bietti2019inductive,allen2019can} and Transformers~\cite{xu2021vitae,chrupala2018symbolic}. Inductive biases can be generally grouped into two camps: 1) Equivariance: the feature representation should faithfully preserve all the data semantics~\cite{cohen2019gauge,cohen2016group,lenssen2018group}. 
2) Invariance: generalization is about learning to be invariant to the diverse environments~\cite{wang2021self,bardes2021vicreg}. Common practical examples are the pooling/striding in CNN~\cite{kayhan2020translation}, dropout~\cite{helmbold2015inductive}, denoising autoencoder~\cite{jo2021rethinking}, batch normalization~\cite{daneshmand2021batch}, and data augmentations~\cite{bouchacourt2021grounding,chen2020group}.

\noindent\textbf{Data-Efficient Learning}. 
Most existing works re-use existing datasets~\cite{zhu2021class,castro2018end} and  synthesize artificial training data~\cite{lahiri2021lipsync3d,cubuk2020randaugment}. We work is more related to those that overcome the data dependency by adding prior knowledge to deep nets~\cite{battaglia2018relational,gondal2019transfer}. Note that data-efficient learning is more general than the popular setting of few-shot learning~\cite{sung2018learning,snell2017prototypical,sun2019meta} which still requires external large pre-training data as initialization or meta-learning. In this work, we offer a theoretical analysis for the difference between learning from insufficient and sufficient data, by posing it in an OOD generalization problem.

\noindent\textbf{OOD Generalization}. 
Conventional machine generalization is based on the Independent and Identically Distributed (IID) assumption for training and testing data~\cite{vapnik1992principles}. However, this assumption is often challenged in practice---the Out-of-Distribution (OOD) problem degrades the generalization significantly~\cite{hendrycks2019benchmarking,zhang2022towards,recht2019imagenet}. Most existing works can be framed into the stable causal effect pursuit~\cite{pfister2019invariant,wang2021causal,jung2020learning} or finding an invariant feature subspace~\cite{pan2010domain,you2019universal}. 
Recently, Invariant Risk Minimization (IRM) takes a different optimization strategy such as convergence speed regularization~\cite{arjovsky2019invariant,krueger2020out} and game theory~\cite{ahuja2020invariant}. Our proposed class-wise IRM makes it more practical by relaxing the restrictions on needing environment annotation.

\section{Justifications of the Two Inductive Biases}

As we discussed in Section~\ref{sec:intro}, given an image $X=x$ with label $Y=y$, our goal is to extract the intrinsic class feature $\phi(x)$ invariant to the environmental changes $z\in Z$.
Specifically, $Z$ is defined as all the class-agnostic items in the task of interest. For example, spatial location is the intrinsic class feature in object detection task, but an environmental feature in image classification. 
This goal can be achieved by using the interventional Empirical Risk Minimization (ERM)~\cite{jung2020learning}. It replaces the observational distribution $P(Y|X)$ with the interventional distribution $P(Y|do(X))$, which removes the environmental effects from the prediction of $Y$, making $Y=y$ only affected by $X=x$~\cite{pearl2009causality}. The interventional empirical risk $\mathcal{R}$ with classifier $f$ can be written as (See Appendix for the detailed derivation): 
\begin{small}
\begin{equation}
\begin{aligned}
    \mathcal{R} &= \mathbb{E}_{x\sim P(X), y\sim P(Y|do(X))} \mathcal{L}(f(\phi(x)), y)\\
    &= \sum_{x}\sum_{y}\sum_{z} \mathcal{L}(f(\phi(x)), y) P(y|x,z) P(z) P(x),
    \label{eq:intervention}
\end{aligned}
\end{equation}
\end{small}
where $\mathcal{L}(f(\phi(x)), y)$ is the standard cross-entropy loss. Note that Eq.~\eqref{eq:intervention} is hard to implement since the environment $Z$ is unobserved in general.

When the training data is sufficient, $X$ can be almost observed in any environment $Z$,  leading to the approximate independence of $Z$ and $X$, \ie, $P(Z|X)\approx P(Z)$. 
Then $\mathcal{R}$ in Eq.~\eqref{eq:intervention} approaches to the conventional ERM $\hat{\mathcal{R}}$:
\begin{small}
\begin{equation}
    \mathcal{R} \approx  \hat{\mathcal{R}}=\sum_{x}\sum_{y} \mathcal{L}(f(\phi(x)), y) P(y|x) P(x)
    = \mathbb{E}_{(x,y)\sim P(X,Y)} \mathcal{L}(f(\phi(x)), y),
    \label{eq:intervention_suff}
\end{equation}
\end{small}

\begin{figure}[t]
\captionsetup{font=footnotesize,labelfont=footnotesize}
    \centering
    \includegraphics[width=.95\textwidth]{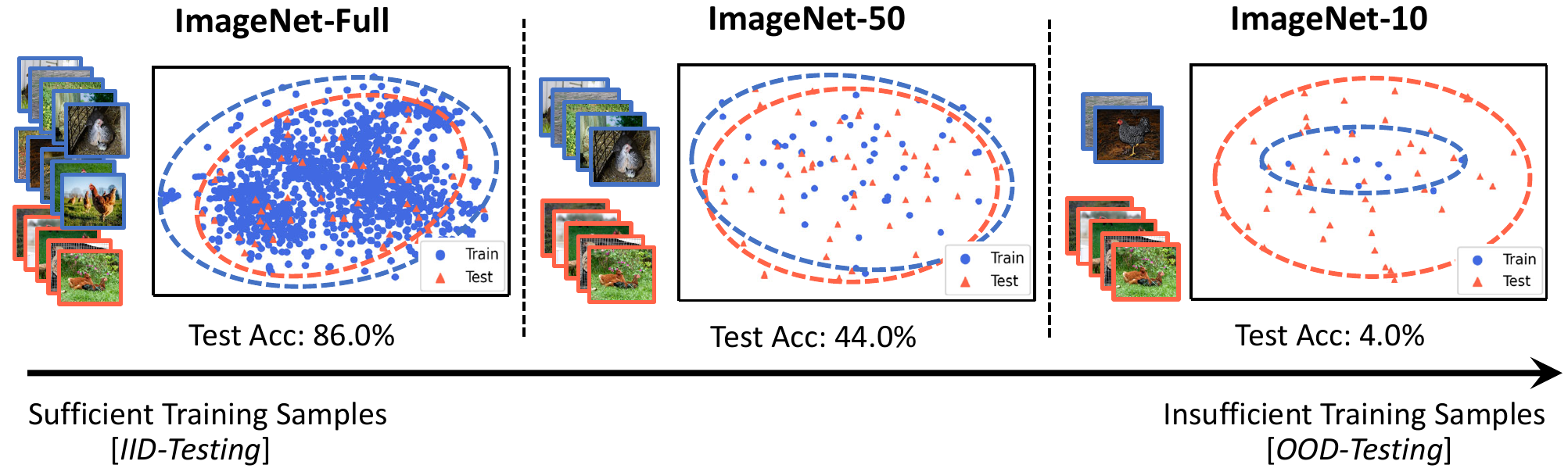}
    \vspace{-3mm}
    \caption{The t-SNE~\cite{van2008visualizing} data visualization of class ``hen'' on different-scale ImageNet dataset using CLIP~\cite{clip} pretrained feature extractor. Blue dot and orange triangle represent training and testing samples, respectively. The testing accuracy is evaluated by ResNet-50~\cite{he2016deep} trained from scratch on each dataset. See Appendix for details.}
    \label{fig:justification_full_low}
\end{figure}

\subsection{Model Deficiency in Data Insufficiency}\label{sec:3_1}
 However, when the training data is insufficient, $P(Z|X)$ is no longer approximate to $P(Z)$ and thus $\hat{\mathcal{R}}\not\approx \mathcal{R}$. For example, $P(Z=\inlineimg{inline_figures/white-inline}|\inlineimg{inline_figures/swan-inline})>P(Z=\inlineimg{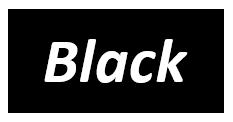}|\inlineimg{inline_figures/swan-inline})$. Then, as we discussed in Section~\ref{sec:intro}, some simple environmental semantics $Z$, \eg, $Z = \inlineimg{inline_figures/white-inline}$, are more likely dominant in minimizing $\hat{\mathcal{R}}$ due to $P(y|x) = P(y|x,z)P(z|x)$ in Eq.~\eqref{eq:intervention_suff}, resulting the learned $\phi$ that mainly captures the dominant environment but missing the intrinsic class feature. Empirical results in Fig.~\ref{fig:justification_full_low} also support such analysis.
We show the ImageNet classification results of class \texttt{hen} using various training sizes. We can observe that with the decreasing of training samples, the accuracy degrades significantly, from 86.0\% to 4.0\%. After all, when the training size is infinite, any testing data is a subset of training.

\subsection{Inductive Bias I: Equivariant Feature Learning}\label{sec:3_2}
To win back the missing intrinsic class feature, we impose the contrastive-based self-supervised learning (SSL) techniques~\cite{chen2020simple,he2019moco,van2018representation}, without the need for any external data, to achieve the equivariance.  In this paper, we follow the definition and implementation in~\cite{wang2021self} to achieve sample-equivariant by using contrastive learning, \ie, different samples should be respectively mapped to different features.
Given an image $x$, the data augmentation of $x$ constitute the positive example $x^{+}$, whereas augmentations of other images constitute $N$ negatives $x^{-}$.
The key of contrastive loss is to map positive samples closer, while pushing apart negative ones in the feature space:
\begin{small}
\begin{equation}
    \mathbb{E}_{x, x^{+}, \{x_{i}^{-}\}_{i=1}^{N}}\left[-\log \frac{\mathrm{exp}(\phi(x)^\mathrm{T} \phi(x^{+}))}{\mathrm{exp}(\phi(x)^\mathrm{T} \phi(x^{+})) + \sum_{i=1}^{N} \mathrm{exp}(\phi(x)^\mathrm{T} \phi(x_i^{-}))}\right].
    \label{eq:contrastive}
\end{equation}
\end{small}
Note that we are open to any SSL choice, which is investigated in Section~\ref{sec:exp}. 

\begin{figure}[t]
    \centering
    \footnotesize
    \includegraphics[width=.99\textwidth]{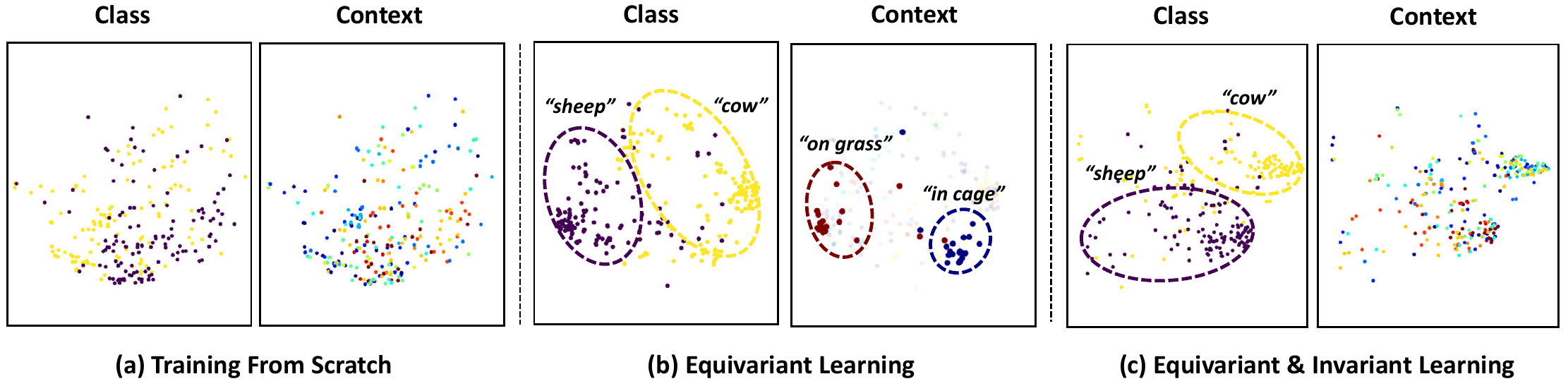}
    \caption{The t-SNE~\cite{van2008visualizing} visualization of learned features \textit{w.r.t} both class and context annotations on NICO dataset with (a) training from scratch; (b) equivariant learning; and (c) equivariant \& invariant learning.}
    \vspace{-3mm}
    \label{fig:justification_equivirance}
\end{figure}

We visualize the features learned by training from scratch and utilizing the equivariance inductive bias on NICO~\cite{he2021towards} dataset with both class and context annotations. In Fig.~\ref{fig:justification_equivirance} (a), it is obvious that there is no clear boundary to distinguish the semantics of class and context in the feature space, while in Fig.~\ref{fig:justification_equivirance} (b), the features are well clustered corresponding to both class and context.

\subsection{Inductive Bias II: Invariant Risk Minimization}\label{sec:3_3}
Although the equivariance inductive bias preserves all the features, the \texttt{swan} model may still be confounded by the ``white'' feature during the downstream fine-tuning, causing $\hat{\mathcal{R}} \neq \mathcal{R}$. To mitigate such shortcut bias, a straightforward solution is to use Inverse Probability Weighting (IPW)~\cite{austin2011introduction,imbens2015causal,little2019statistical} (also known as reweighting~\cite{bahng2020learning,nam2020learning,lee2021learning}) to down weight the overwhelmed ``white'' feature in \texttt{swan}. However, they must follow the positivity assumption~\cite{hernan2010causal}, \ie, all the environmental semantics $Z$ should exist in each class. However, when the training data is insufficient, such assumption no longer holds. For example, how do you down weight ``white'' over ``black'' if there is even no ``black swan'' sample?

Recently, Invariant Risk Minimization (IRM)~\cite{arjovsky2019invariant,krueger2020out} resolves the non-positivity issue by imposing the invariance inductive bias to directly remove the effect of environmental semantics $Z$. Specifically, IRM first splits the training data into multiple environments $e\in \mathcal{E}$. Then, it regularizes $\phi$ to be \emph{equally} optimal in different splits, \ie, invariant across environments:
\begin{small}
\begin{equation}
    \sum_{e} \mathcal{L}_{e} (w^\mathrm{T} \phi(x), y) + \lambda \|\nabla_{w=1}\mathcal{L}_{e} (w^\mathrm{T} \phi(x), y)\|_2^2,
    \label{eq:irm}
\end{equation}
\end{small}
where $\lambda$ is trade-off hyper-parameter, $w$ stands for a dummy classifier~\cite{arjovsky2019invariant} to calculate the gradient penalty across splits---though different environments may induce different losses, the feature $\phi$ must regularize them optimal at the same time (the lower gradient the better) in the same way (by using the common dummy classifier). Note that each environment should contain a unique mode of environmental feature distribution~\cite{arjovsky2019invariant,creager2021environment,ahuja2020empirical}: suppose that we have $k$ environmental features that are distributed as $\{p_1, p_2, ..., p_k\}$.
If $p^{e_1}_i\neq p^{e_2}_i$, $i=1$ to $k$, IRM under the two environments will remove all the $k$ features---the keeping of any one will be penalized by the second term of Eq.~\eqref{eq:irm}.

Conventional IRM requires the environment annotations, which are generally impossible in practice. To this end, we propose a novel class-wise IRM to regularize the invariance within each class, without the need for environment supervision.
We show the qualitative results of imposing such invariance inductive bias in Fig.~\ref{fig:justification_equivirance}~(c). Compared to Fig.~\ref{fig:justification_equivirance}~(b), we can observe that after applying our proposed class-wise IRM, the equivariance of intrinsic class features are reserved with well-clustered data points while the context labels are no longer responsive---the environment features are removed.

\section{Our \algname Algorithm}\label{sec:4}
\label{sec:approach}

\begin{figure}[t]
\captionsetup{font=footnotesize,labelfont=footnotesize}
    \centering
    \includegraphics[width=.98\textwidth]{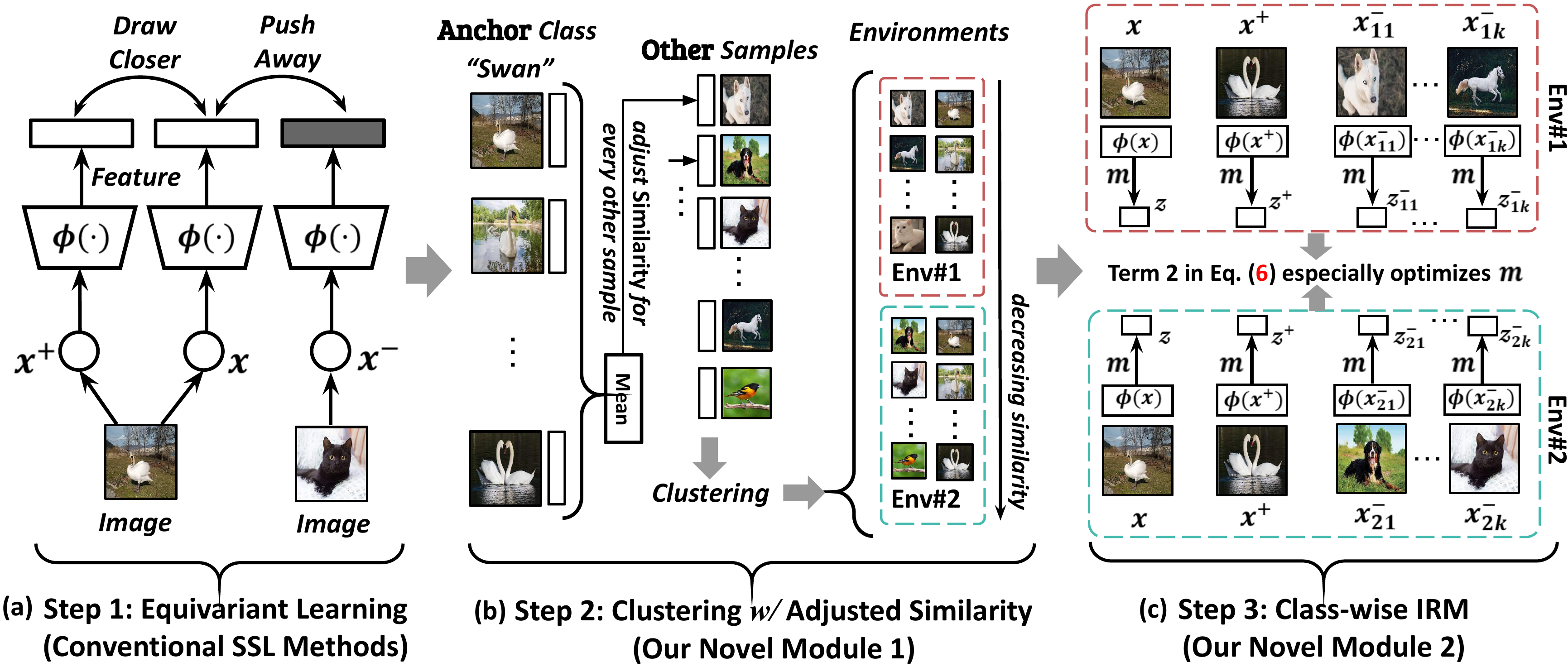}
    \caption{The flowchart of our proposed \algname with 3 steps. Rectangle with shading denotes the feature and $\mathcal{E}_j$ represents the generated environment of class $j$. $x_{1k}^{-}$ and $x_{2k}^{-}$ in (c) are the $k$-th negative samples of subset $e_1$ and $e_2$, respectively. We highlight that class-wise IRM optimizes the mask layer $m$ (and an extra MLP $g$) without gradients flowing back to the feature extract $\phi$.}
    \vspace{-1mm}
    \label{fig:framework}
\end{figure}

Fig.~\ref{fig:framework} depicts the computing flow of \algname. In the following, we elaborate each of its components.

\noindent\textbf{Input:} Insufficient training samples denoted as the pairs $\{(x,y)\}$ of an image $x$ and its label $y$.

\noindent\textbf{Output:} Robust classification model $f\cdot\phi$ with intrinsic class feature $\phi(x)$ and unbiased classifier $f(\phi(x))$.

\noindent\textbf{Step 1: Equivariant Learning via SSL.}
As introduced in Section~\ref{fig:justification_equivirance}, a wide range of SSL pretext tasks are sufficient for encoding the sample-equivariance. For fair comparison with other methods in VIPriors challenge dataset~\cite{bruintjes2021vipriors}, we use MoCo-v2~\cite{he2019moco,chen2020mocov2}, Simsiam~\cite{chen2020exploring}, and IP-IRM~\cite{wang2021self} to learn $\phi$ in Fig~\ref{fig:framework}~(a). We leave the results based on the most recent MAE~\cite{he2021masked} in Appendix.

\noindent\textbf{Step 2: Environment Construction based on Adjusted Similarity.}
Now we are ready to use IRM to remove the environmental features in $\phi$. Yet, conventional IRM does not apply as we do not have environment annotations. So, this step aims to automatically construct environments $\mathcal{E}$. However, it is extremely challenging to identify the combinatorial number of unique environmental modes---improper environmental split may contain shared modes, which cannot be removed. To this end, we propose an efficient \emph{class-wise} approximation that seeks \emph{two} environments \textit{w.r.t.} each class. Our key motivation is that, for insufficient training data, the environmental variance within each class is relatively simple and thus we assume that it is single-modal. 
Therefore, as shown in Fig.~\ref{fig:framework}~(b), we propose to use each class (we call \texttt{anchor} class) as an anchor environmental mode to split the samples of
the rest of the classes (we call \texttt{other} classes)
into two groups: similar to the \texttt{anchor} or not. As a result, for $C$ classes, we will have totally $2C$ approximately unique environments. Intuitively, this class-wise strategy can effectively remove the severely dominant context bias in a class. For example, if all \texttt{swan} samples are ``white'', the ``white'' feature can still be identified as a non-discriminative color feature, thanks to the ``black'' and ``white'' samples of \texttt{dog} class.

\begin{figure}[t]
\captionsetup{font=footnotesize,labelfont=footnotesize}
    \centering
    \includegraphics[width=1.0\textwidth]{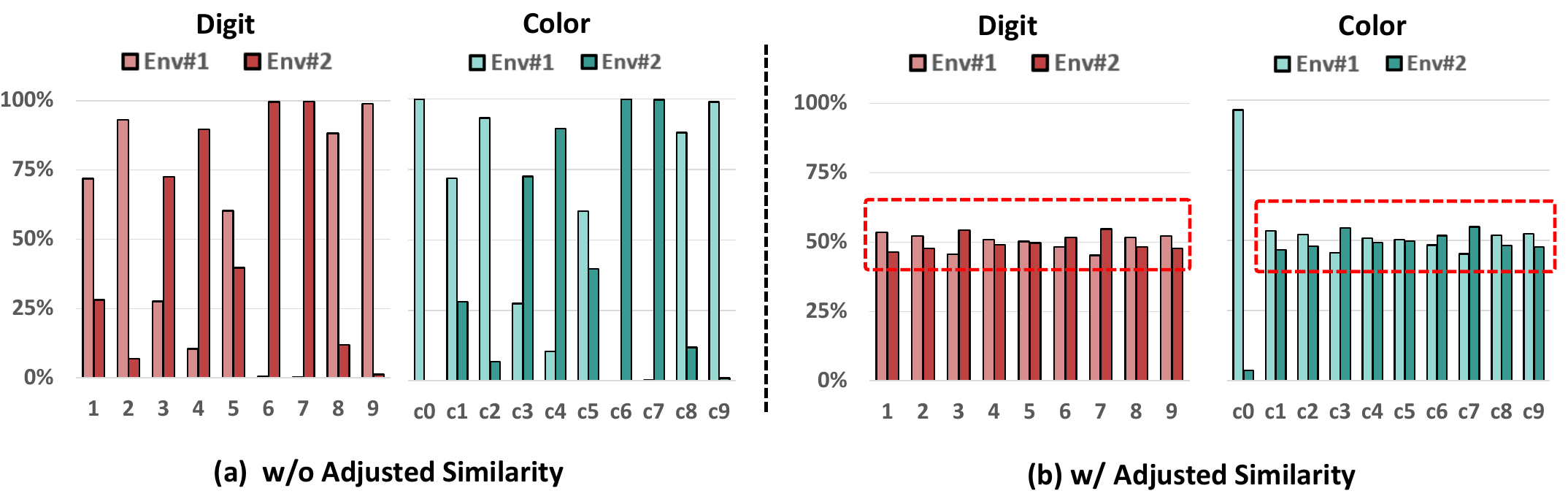}
    \caption{The obtained environments $\mathcal{E}_0$  for an example \texttt{anchor} class \texttt{0} on the Colored MNIST~\cite{nam2020learning},
    by using (a) the vanilla cosine similarity and (b) our adjusted similarity. On X-axis, \texttt{1}-\texttt{9} are \texttt{other} digit classes, and \texttt{c0}-\texttt{c9} denote $10$ colors used to create this color-bias dataset. On Y-axis, the percentage point denotes the proportion of a digit (or a color) grouped into a specific environment.}
    \vspace{-2mm}
    \label{fig:cmnist_cluster}
\end{figure}

For each \texttt{anchor} class containing $l$ images, environment Env\#1 contains these $l$ samples as positive and the ``similar'' samples from \texttt{other} classes as negative; environment Env\#2 contains the same positive samples while the ``dissimilar'' samples from \texttt{other} classes as negative. 
A straightforward way to define the ``similarity'' between two samples is to use cosine similarity.
We compute the cosine similarity between the pair images sampled from \texttt{anchor} class and \texttt{other} classes, respectively.
We get the matrix $\mathbf{S}\in \mathbb{R}^{l\times n}$, where $n$ is the number of images in \texttt{other} classes. Then, we average this matrix along the axis of \texttt{anchor} class, as in the pseudocode: $\mathbf{s}^+ = \textrm{mean}(\mathbf{S}, \mathrm{dim=0})$.
After ranking $\mathbf{s}^+$, it is easy to get ``similar'' samples (corresponding to higher half values in $\mathbf{s}^+$) grouped in Env$\#$1 and ``dissimilar'' samples (corresponding to lower half values in $\mathbf{s}^+$) grouped in Env$\#$2. 
It is an even split.
Fig.~\ref{fig:cmnist_cluster}~(a) shows 
the resultant environments for \texttt{anchor} class \texttt{0} on the Colored MNIST\footnote[3]{\scriptsize{It is modified from MNIST dataset~\cite{lecun2010mnist} by injecting \textit{color} bias on each digit (class).
The non-bias ratio is $0.5\%$, \eg, $99.5\%$ samples of \texttt{0} are in red and only $0.5\%$ in uniform colors.}}~\cite{nam2020learning} using the above straightforward cosine similarity. We can see that the digit classes distribute differently in Env$\#$1 and Env$\#$2, indicating that the difference of the two environments also include class information, which will be disastrously removed after applying IRM.

To this end, we propose a similarity adjustment method.
It is to adjust every \underline{sample}-to-class similarity by subtracting a \underline{class}-to-class similarity, where the \underline{sample} belongs to the \underline{class}. 
First, we calculate the class-to-class similarity $\bar{s}_i$ between the $i$-th ($i = 1, ..., C-1$) \texttt{other} class and the \texttt{anchor} class: $\bar{s}_i=\mathrm{mean}(\mathbf{s}^+[a_i:b_i])$, where we assume that the image index range of the $i$-th \texttt{other} class is $[a_i:1:b_i]$. Such similarity can be viewed as a purer ``class effect'' to be removed from the total effect of both class and environment---only ``environment effect'' is then left. Therefore, for any sample $x^j$ from the $i$-th \texttt{other} class, its adjusted similarity to the \texttt{anchor} class is: $s = \mathbf{s}^+[j] -\bar{s}_i$. Using this similarity, we obtain new environments and show statistics in Fig.~\ref{fig:cmnist_cluster}~(b). It is impressive that the biased color of \texttt{anchor} class \texttt{0} (\ie, the $0$-th color $c0$ or \texttt{red}) varies between Env$\#$1 and Env$\#$2, but the classes and other colors (red dashed boxes) distribute almost uniformly in these two environments. It means the effects of class and environment are disentangled.

\noindent\textbf{Step 3: Class-wise Invariant Risk Minimization}.
With the automatically constructed environments, we are ready to remove the environmental feature from $\phi$. In particular, we propose a class-wise IRM based on the contrastive objective, which is defined as follows. As shown in Fig.~\ref{fig:framework}~(c), given a training image $x$ in environment $e$ of class $i$, we use a learnable vector mask layer $m$ multiplied on $\phi(x)$ to select the invariant feature.
Then, we follow~\cite{chen2020simple} to build a projection layer $g(\cdot)$ to obtain $\bm{z}=g(m \circ \phi(x))$ for contrastive supervision, where $g$ is a one-hidden-layer MLP with ReLU activation and $\circ$ denotes element-wise production. For each \texttt{anchor} class $k$, we define an environment-based supervised contrastive loss~\cite{khosla2020supervised}. It is different from the traditional self-supervised contrastive loss. Specifically, our loss is computed within each environment $e\in\mathcal{E}_k$. 
We take the representations of \texttt{anchor} class samples (in $e$) 
as positives $\bm{z}^+$, and the representations of \texttt{other} class samples (in $e$)  as negatives $\bm{z}^-$, and we have:
\begin{small}
\begin{equation}
    \ell(e\in\mathcal{E}_k, w=1) = \sum_{\bm{z}\in e} \frac{1}{N^+} \sum_{\bm{z}^{+}\in e}\left[-\log \frac{\mathrm{exp}(\bm{z}^\mathrm{T} \bm{z}^{+} \cdot w)}{\mathrm{exp}(\bm{z}^\mathrm{T} \bm{z}^{+}\cdot w) + \sum_{\bm{z}^{-}\in e} \mathrm{exp}(\bm{z}^\mathrm{T} \bm{z}^{-}\cdot w)}\right],
    \label{eq:contrastive_sup}
\end{equation}
\end{small}
\noindent
where $N^+$ denotes the number of the positive samples in the current minibatch and $w=1$ is a ``dummy'' classifier to calculate the gradient penalty term~\cite{arjovsky2019invariant}.
Therefore, the proposed class-wise IRM loss\footnote[4]{\scriptsize{Please note that in implementation, we adopt an advanced version~\cite{krueger2020out} of IRM. Please check appendix for details.}} is:
\begin{equation}
    \mathcal{L}_k = \sum\nolimits_{e\in\mathcal{E}_k} \ell (e, w=1) + \lambda \|\nabla_{w=1}\ell (e, w=1)\|_2^2,
    \label{eq:supcont_irm}
\end{equation}
where $\lambda$ is the trade-off hyper-parameter. The overall training objective is the combination of minimizing a conventional cross entropy $\mathcal{L}_{ce}$ and the class-wise IRM regularization $\mathcal{L}_k$:
\begin{equation}
    \min_{f,g,m,\phi} \mathcal{L}_{ce}(f,m,\phi) + \sum\nolimits_{k=1}^{C} \mathcal{L}_k(g,m),
    \label{eq:overall_loss}
\end{equation}
where we use $f(m\circ\phi(x))$ for inference. It is worth noting that each loss trains a different set of parameters---$\phi$ is frozen during the class-wise IRM penalty update. 
As the equivariance of $\phi$ is only guaranteed by SSL pretraining, compared to the expensive SSL equivariance regularization in training~\cite{wang2021self}, our frozen strategy is more efficient to mitigate the adversary effect introduced by the invariance bias, which may however discard equivariant features to achieve invariance. We investigate this phenomenon empirically in Section~\ref{sec:ablation_study}.

\section{Experiments}
\label{sec:exp}

\subsection{Datasets and Settings}

\noindent\textbf{VIPriors}~\cite{bruintjes2021vipriors}
dataset is proposed in VIPrior challenge~\cite{bruintjes2021vipriors} for data-efficient learning. It contains the same 1,000 classes as in ImageNet~\cite{deng2009imagenet}, and also follows the same splits of \texttt{train}, \texttt{val} and \texttt{test} data. In all splits, each class contains only $50$ images, so the total number of samples in the dataset is $150k$. Some related works~\cite{barz2021strong,liu2020diversification} used the merged set (of \texttt{train} and \texttt{val}) to train the model. We argue that this to some extent violates the protocol of data-efficient learning---using insufficient training data. In this work, our \algname models as well as comparing models are trained on the standard \texttt{train} set and evaluated on \texttt{val} and \texttt{test} sets. In addition, we propose two more challenging settings to evaluate the models: \textbf{VIPriors-20} and \textbf{VIPriors-10}. The only difference with VIPriors is they have $20$ and $10$ images per class in their \texttt{train} sets, respectively. There is no change on \texttt{val} and \texttt{test} sets.
We thus call the original \textbf{VIPriors-50}. \noindent\textbf{NICO}~\cite{he2021towards} is a real-world image dataset proposed for evaluating OOD methods. The key insight of NICO is to provide image labels as well as context labels (annotated by human). On this dataset, it is convenient to ``shift'' the distribution of the class by ``adjusting'' the proportions of specific contexts. In our experiments, we follow the ``adjusting'' settings in the related work~\cite{wang2021causal}. Specifically, this is a challenging OOD setting using the NICO animal set. It mixes three difficulties: 1) Long-Tailed; 2) Zero-Shot and 3) Orthogonal.
\noindent\textbf{ImageNet100}~\cite{tian2019contrastive} is a subset of original ImageNet~\cite{deng2009imagenet} with $100$ classes and 1$k$ images per class. Different with previous OOD datasets, ImageNet100 is to evaluate the performances of our~\algname and comparison methods in sufficient training data settings.

\begin{table}[t]
\captionsetup{font=footnotesize,labelfont=footnotesize}
\centering
\caption{Recognition accuracies (\%) on the VIPriors-50, -20, -10, NICO and ImageNet-100 (IN-100) datasets. ``Aug.'' represents augmentation. Note that due to the effectiveness of ``Random Aug.'', we set it as a default configuration for the methods trained from SSL. Our results are highlighted.}
\scalebox{0.78}{
\begin{tabular}{p{0.7cm}<{\centering}p{3.1cm}ccccccccccccccc}
\toprule\toprule
\multicolumn{2}{c}{\multirow{2}{*}{\large{Model}}} & \multicolumn{2}{c}{VIPriors-50~\cite{bruintjes2021vipriors}} &~ & \multicolumn{2}{c}{~~VIPriors-20~~} &~ & \multicolumn{2}{c}{~~VIPriors-10~~} &~ & \multicolumn{2}{c}{~~~NICO~\cite{he2021towards}~~~} &~ & 
\multicolumn{2}{c}{IN-100~\cite{tian2019contrastive}} &~ \\ \cmidrule(lr){3-4}\cmidrule(lr){6-7}\cmidrule(lr){9-10}\cmidrule(lr){12-13} \cmidrule(lr){14-16} 
\multicolumn{2}{c}{} & \multicolumn{1}{c}{Val} & \multicolumn{1}{c}{Test} &~ & \multicolumn{1}{c}{Val} & \multicolumn{1}{c}{Test} &~ & \multicolumn{1}{c}{Val} & \multicolumn{1}{c}{Test} &~ & \multicolumn{1}{c}{Val} & \multicolumn{1}{c}{Test} &~ & \multicolumn{2}{c}{Val} &  \\ \hline
\multirow{10}{*}{\rotatebox{90}{\small{Train from Scratch}}} & Baseline & 32.30 & 30.60 &~ & 13.13 & 12.39 &~ & 5.02 & 4.59 &~ & 43.08 & 40.77 &~ & & 83.56\\ \cline{2-16} 
 & \textit{Augmentation} &  &  &~ &  &  &~ &  &  &~ &  &  &~ \\
 & Stronger Aug.~\cite{chen2020simple} & 36.60 & 34.72 &~ & 16.17 & 15.21 &~ & 3.49 & 3.26 &~ & 42.31 & 43.31 &~ & & 83.72\\
 & Random Aug.~\cite{cubuk2020randaugment} & 41.09 & 39.18 &~ & 16.71 & 16.03 &~ & 3.88 & 4.01 &~ & 45.15 & 44.92 &~ & & 85.12\\
 & Mixup~\cite{zhang2017mixup} & 34.66 & 32.75 &~ & 13.35 & 12.69 &~ & 2.47 & 2.31 &~ & 40.54 & 38.77 &~ & & 84.52\\
 & Label smoothing~\cite{muller2019does} & 33.77 & 31.87 &~ & 12.71 & 12.05 &~ & 4.76 & 4.43 &~ & 39.77 & 38.15 &~ & & 85.22\\ \cline{2-16} 
 & \textit{Debias Learning} &  &  &~ &  &  &~ &  &  &~ &  &  &~ \\
 & Lff~\cite{nam2020learning} & 35.04 & 33.29 &~ & 13.26 & 12.58 &~ & 5.20 & 4.79 &~ & 41.62  & 42.54 &~ & & 83.74\\
 & Augment Feat.~\cite{lee2021learning} & 35.41 & 33.63 &~ & 13.62 & 12.97 &~ & 3.43 & 3.12 &~ & 42.31 & 43.27 &~ & & 83.88\\
 & CaaM~\cite{wang2021causal} & 36.13 & 34.24 &~ & 14.68 & 13.99 &~ & 4.88 & 4.63 &~ & 46.38 & 46.62 &~ & & 84.56 \\ \hline
\multirow{6}{*}{\rotatebox{90}{\small{Train from SSL}}} 
& MoCo-v2~\cite{chen2020mocov2} & 49.47 & 46.98 &~ & 30.76 & 28.83 &~ & 18.40 & 16.97 &~ & 46.45 & 45.70 &~ & & 86.30\\ 
 & ~~+\algname (Ours) & \cellcolor{mygray}{\textbf{54.21}} & \cellcolor{mygray}{\bf 52.09} & \cellcolor{mygray}{~} & \cellcolor{mygray}{\textbf{38.30}} & \cellcolor{mygray}{\bf 36.66} & \cellcolor{mygray}{~} & \cellcolor{mygray}{\textbf{26.70}} & \cellcolor{mygray}{\bf 25.20} &\cellcolor{mygray}{~} & \cellcolor{mygray}{\textbf{52.55}} & \cellcolor{mygray}{\bf 51.51} &\cellcolor{mygray}{~} &\cellcolor{mygray} &\cellcolor{mygray} \textbf{88.38}\\ \cline{2-16} 
 & SimSiam~\cite{chen2020exploring} & 42.69 & 40.75 &~ & 22.09 & 21.15 &~ & 6.84 & 6.68 &~ & 41.27 & 42.68 &~ & & 85.28\\ 
  & ~~+\algname (Ours) & \cellcolor{mygray}{\textbf{52.55}} & \cellcolor{mygray}{\bf 50.36} &\cellcolor{mygray}{~} & \cellcolor{mygray}{\textbf{37.29}} & \cellcolor{mygray}{\bf 35.65} &\cellcolor{mygray}{~} & \cellcolor{mygray}{\textbf{24.74}} & \cellcolor{mygray}{\bf 23.33} &\cellcolor{mygray}{~} & \cellcolor{mygray}{\textbf{45.67}} & \cellcolor{mygray}{\bf 44.77} &\cellcolor{mygray}{~} &\cellcolor{mygray} &\cellcolor{mygray} \textbf{86.80}\\ \cline{2-16}
 & IP-IRM~\cite{wang2021self} & 51.45 & 48.90 &~ & 38.91 & 36.26 &~ & 29.94 & 27.88 &~ & 63.60 & 60.26 &~ & & 86.94\\ 
 & ~~+\algname (Ours) & \cellcolor{mygray}{\textbf{54.58}} & \cellcolor{mygray}{\bf 52.27} &\cellcolor{mygray}{~} & \cellcolor{mygray}{\textbf{41.53}} & \cellcolor{mygray}{\bf 39.21} &\cellcolor{mygray}{~} & \cellcolor{mygray}{\textbf{32.70}} & \cellcolor{mygray}{\bf 30.36} &\cellcolor{mygray}{~} & \cellcolor{mygray}{\textbf{66.07}} & \cellcolor{mygray}{\bf 64.14} &\cellcolor{mygray}{~} &\cellcolor{mygray} &\cellcolor{mygray} \textbf{87.78}\\ \bottomrule\bottomrule
\end{tabular}}
\label{tab:overall_results}
\end{table}

\subsection{Implementation Details}

We adopted ResNet-50/-18 as model backbones for VIPriors/ImageNet100 and NICO datasets, respectively. 
We trained the model with 100 epochs for ``training from scratch'' methods. We initialized the learning rate as $0.1$ and decreased it by $10$ times at the $60$-th and $80$-th epochs. We used SGD optimizer and the batch size was set as 256. For equivariant learning (\ie, SSL), we utilized MoCo-v2~\cite{chen2020mocov2}, SimSiam~\cite{chen2020exploring} and IP-IRM~\cite{wang2021self} to train the model for 800 epochs without using external data, using their default hyper-parameters. We pretrain the model for 200 epochs on ImageNet100 dataset. 
Then for downstream fine-tuning, We used SGD optimizer and set batch size as 128. We set epochs as 50, initialized learning rate as 0.05, and decreased it at the $30$-th and $40$-th epochs. 
Please check appendix for more implementation details.
Below we introduce our baselines including augmentation-based methods, debiased learning methods and domain generalization (DG) methods.

\noindent\textbf{Augmentation-based Methods} are quite simple yet effective techniques in the VIPriors challenges as well as for the task of data-efficient learning. 
We chose four top-performing methods in this category to compare with: stronger augmentation~\cite{chen2020simple}, random augmentation\cite{cubuk2020randaugment}, mixup~\cite{zhang2017mixup} and label smoothing~\cite{muller2019does}.

\noindent\textbf{Debias Learning Methods}. 
Data-efficient learning can be regarded as a task for OOD.
We thus compared our \algname with three state-of-the-art (SOTA) debiased learning methods: Lff~\cite{nam2020learning}, Augment Feat.~\cite{lee2021learning} and CaaM~\cite{wang2021causal}.

\noindent\textbf{Domain Generalization Methods}.
Domain Generalization (DG) task also tackles the OOD generalization problem, but requires sufficient domain samples and full ImageNet pretraining.
In this paper, we select three SOTA DG approaches (SD~\cite{pezeshki2021gradient}, SelfReg~\cite{kim2021selfreg} and SagNet~\cite{nam2021reducing}) for comparison. These methods do not require domain labels which share the same setting as ours.

\subsection{Comparing to SOTAs}

Table~\ref{tab:overall_results} 
shows the overall results comparing to baselines on VIPriors-50, -20, -10, NICO and ImageNet100 datasets.
Our \algname achieves the best performance across all settings.
In addition, we have another four observations. 1) Incorporating SSL pretraining, vanilla fine-tuning can achieve much higher accuracy than all the methods of ``training from scratch''. 
This validates the efficiency of the equivariance inductive bias (learned by SSL) for Etackling the challenge of lacking training data.
2) When decreasing the training size of VIPriors from 50 to 10 images per class, the comparison methods of training from scratch cannot bring performance boosting even hurt the performance. This is because the extremely insufficent data cannot support to establish an equivariant representation, not mention to process samples with harder augmentations.
3) Interestingly, compared to SSL methods, we can see that the improvement margins by our method are larger in the more challenging VIPriors-10, \eg, $8.2\%$ on MoCo-v2 and $16.7\%$ on SimSiam. 
It validates the invariance inductive bias learned by the class-wise IRM (in our \algnamens) helps to disentangle and alleviate the OOD bias effectively.
4) Results on ImageNet100 dataset show the consistent improvements of~\algname due to the additional supervised contrastive loss, indicating the generalizability of our~\algname in a wide range of cases from insufficient to sufficient data.

 \begin{wraptable}{l}{0.55\textwidth}
 \captionsetup{width=0.5\textwidth,font=footnotesize,labelfont=footnotesize,skip=1pt}
\vspace{-0.5cm}
\centering
\caption{Test accuracy (\%) of DG SOTA methods. V-50/-10 denote VIPriors-50/-10.}
\scalebox{0.7}{
\begin{tabular}{p{0.4cm}lccp{0.4cm}lcc}
\toprule \toprule
\multicolumn{1}{l}{} & \multicolumn{1}{c}{Methods} & V-50 & V-10 &  & \multicolumn{1}{c}{Methods} & V-50 & V-10 \\ \midrule
\multirow{6}{*}{\rotatebox{90}{\small{Train from Scratch}}} & Boardline &30.60  &4.59  & \multirow{6}{*}{\rotatebox{90}{\small{Train from SSL}}} & IP-IRM & 48.90 & 26.88 \\ \cmidrule{2-4} \cmidrule{6-8} 
 & SD~\cite{pezeshki2021gradient} & 33.91 & 4.85 &  & ~~+SD~\cite{pezeshki2021gradient} &49.91  &28.01  \\
 & SelfReg~\cite{kim2021selfreg} & 23.85 & 3.64 &  & ~~+SelfReg~\cite{kim2021selfreg} &36.48  &22.75  \\
 & SagNet~\cite{nam2021reducing} & 34.92 & 5.62 &  & ~~+SagNet~\cite{nam2021reducing} &47.82  &26.17  \\ \cmidrule{2-4} \cmidrule{6-8} 
 & \multicolumn{3}{l}{\diagbox[innerwidth = 10.em,height = 3ex]{}{}}  &  & \cellcolor{mygray}~~+\algname & \cellcolor{mygray}\textbf{52.27} & \cellcolor{mygray}\textbf{30.36} \\ \bottomrule \bottomrule
\end{tabular}}
\vspace{-0.4cm}
\label{tab:dg}
\end{wraptable}
In Table~\ref{tab:dg}, we compare our~\algname with DG methods. We try both ``train from scratch'' and ``train from SSL'' to meet the pretraining requirement of DG. We can find that our~\algname outperforms DG methods with large margins, showing the weaknesses of existing OOD methods for handling \emph{insufficient} data.

In Table~\ref{tab:sota}, we compare our \algname with the solutions from other competition teams in the challenge with the same comparable setting: no \texttt{val} is used for training, single model w/o ensemble, similar ResNet50/ResNext50 backbones.
We can observe the best performance is by our method. 
It is worth noting that the competitors Zhao~\etal also used SSL techniques for pretraining. They took the knowledge distillation~\cite{heo2019comprehensive,hinton2015distilling} as their downstream learning method. Our~\algname outperforms their model with a large margin.

\subsection{Ablation Study}
\label{sec:ablation_study}

\noindent\textbf{Q1:} \textit{What are the effects of different components of \algnamens?}

\noindent\textbf{A1:} We traversed different combinations of our proposed three steps to evaluate their effectiveness. The results are shown in Table~\ref{tab:component}. We can draw the following observations: 1) By focusing on the first three rows, we can find that the improvements are relatively marginal without the SSL equivariance pretraining. This is reasonable as the feature similarity cannot reflect the semantics change exactly without the equivariance property, thus affects the environments construction (Step 2) and class-wise IRM (Step 3);
2) The comparison between row 4 to 6 indicates the significance of our proposed similarity adjustment (Step 2). It is clear that the vanilla cosine similarity results in clear performance drops due to the inaccurate environment construction.

\begin{minipage}{\textwidth}
\captionsetup{font=footnotesize,labelfont=footnotesize}
\vspace{2mm}
\begin{minipage}[t]{0.45\textwidth}
    \centering
    \makeatletter\def\@captype{table}\makeatother\caption{Accuracy (\%) comparison with other competition teams (single model w/o ensemble) on the \texttt{val} set of VIPriors-50.} 
\scalebox{0.8}{
\begin{tabular}{ccc}
\toprule\toprule
Team & Backbone & Val Acc \\ \cmidrule(lr){1-1}\cmidrule(lr){2-2}\cmidrule(lr){3-3}
Official Baseline & ResNet50 & 33.16 \\
Zhao~\etal~\cite{zhao2020distilling} & ResNet50 & 44.60 \\
Wang~\etal~\cite{bruintjes2021vipriors} & ResNet50 & 50.70 \\
Sun~\etal~\cite{sun2020visual} & ResNext50 & 51.82 \\
\algname (Ours) & ResNet50 & 54.58 \\ \bottomrule\bottomrule
\end{tabular}}
\label{tab:sota}

\end{minipage}
\hspace{3mm}
\begin{minipage}[t]{0.45\textwidth}
\centering
\makeatletter\def\@captype{table}\makeatother\caption{Evaluation of the effectiveness of our three steps in \algname on VIPriors-20.}\vspace{-1mm}
\scalebox{0.7}{ 
\begin{tabular}{ccccc}
\toprule \toprule
\multicolumn{3}{c}{Components} & \multirow{2}{*}{~~Val~~} & \multirow{2}{*}{~~Test~~} \\ \cline{1-3}
~Step~1~ & ~Step~2~ & ~Step~3~ &  &  \\ \midrule
\XSolidBrush & \XSolidBrush & \XSolidBrush & 13.13 & 12.39 \\
\XSolidBrush & \XSolidBrush & \Checkmark & 13.01 & 12.41 \\
\XSolidBrush & \Checkmark & \Checkmark & 15.69 & 14.17 \\
\Checkmark & \XSolidBrush & \XSolidBrush & 37.61 & 34.88 \\
\Checkmark & \XSolidBrush & \Checkmark & 38.87 & 36.34 \\
\Checkmark & \Checkmark & \Checkmark & 40.15 & 37.78 \\ \bottomrule\bottomrule
\end{tabular}}
\label{tab:component}
\end{minipage}
\end{minipage}
\vspace{2mm}

\noindent\textbf{Q2:} \textit{What is the optimal $\lambda$ for \algnamens? Why does not the class-wise IRM penalty term update feature backbone $\phi$?}

\begin{figure}[t]
\captionsetup{font=footnotesize,labelfont=footnotesize}
    \centering
    \includegraphics[width=0.9\textwidth]{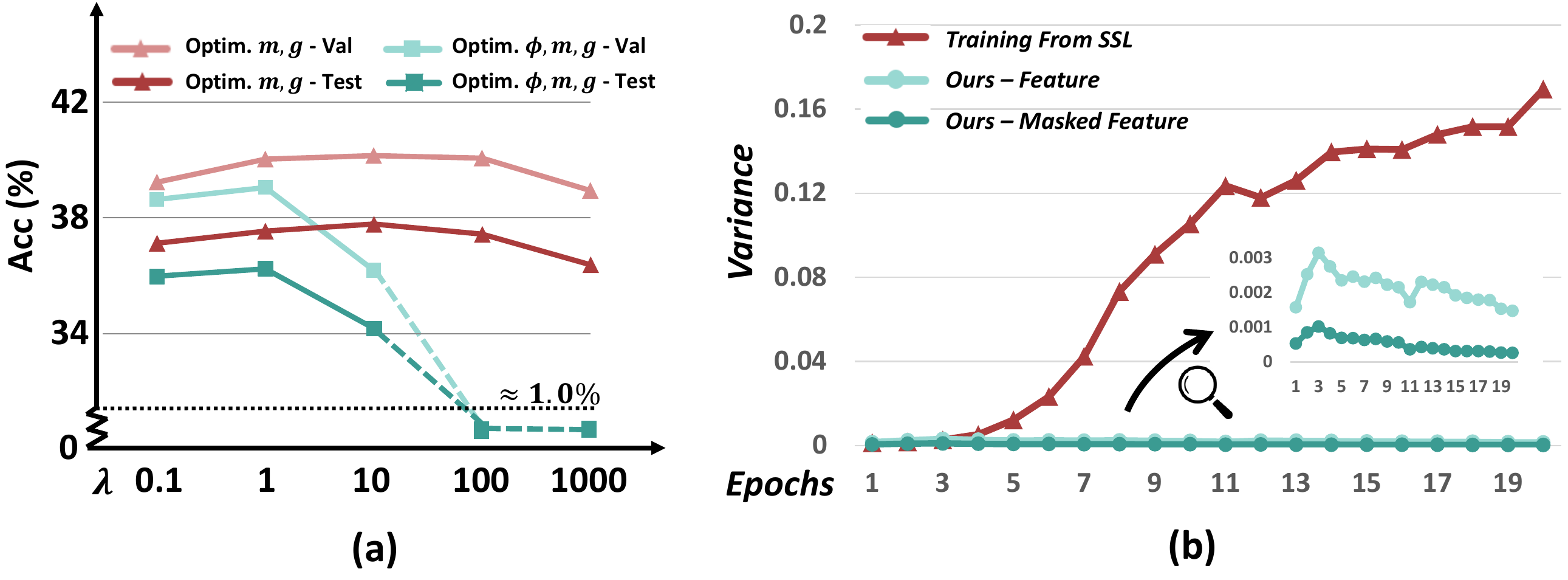}
    \vspace{-3mm}
    \caption{(a) Accuracies (\%) with different optimization schedules and values of $\lambda$ on the VIPriors-20 \texttt{val} and \texttt{test} sets. (b) The intra-class feature variance of training from SSL and our \algname on VIPriors-10 dataet during training process. ``Feature'' and ``Masked Feature'' represent $\phi(x)$ and $m\circ \phi(x)$, respectively.}
    \label{fig:lambda_var}
\end{figure}

\noindent\textbf{A2:} Recall that we highlight such elaborate design in Section~\ref{sec:approach} Step 3. In Fig.~\ref{fig:lambda_var} (a), we evaluate the effect of freezing $\phi$ for Eq.~\eqref{eq:supcont_irm} on VIPriors dataset. First, we can see that setting $\lambda=10$ with freezing $\phi$ can achieve the best validation and test results. Second, when increasing $\lambda$ over $10$, we can observe a sharp performance drops for updating $\phi$, even down to the random guess ($\approx$ 1\%). In contrast, the performances are much more robust with freezing $\phi$ while varying $\lambda$, indicating the non-sensitivity of our $\algname$. This validate the adversary effect of the equivariance and invariance. Updating $\phi$ with large $\lambda$ would destroy the previously learnt equivaraince inductive bias.

\noindent\textbf{Q3:} \textit{Does our \algname achieve invariance with the learned environments $\mathcal{E}$ and the proposed class-wise IRM (\ie, Step 3)?}

\noindent\textbf{A3:} In Fig.~\ref{fig:lambda_var} (b), we calculate the variance of intra-class feature with training from SSL and our \algname on VIPriors-10 data. It represents the feature divergence within the class. We can find that: 1) Compared to our \algname, the variance of training from SSL increases dramatically, indicating that equivariant features are still easily biased to environments without invariance regularizations. 2) The masked feature $m\circ \phi(x)$ of our \algname achieves continuously lower variance than $\phi(x)$, validates the effectiveness of our learnt mask. See Appendix for more visual attention visualizations.

\noindent\textbf{Q4:} \textit{What does the cluster look like for real data with the proposed similarity adjustment (\ie, Step 2)?}

\begin{figure}[t]
\captionsetup{font=footnotesize,labelfont=footnotesize}
    \centering
    \footnotesize
    \includegraphics[width=.99\textwidth]{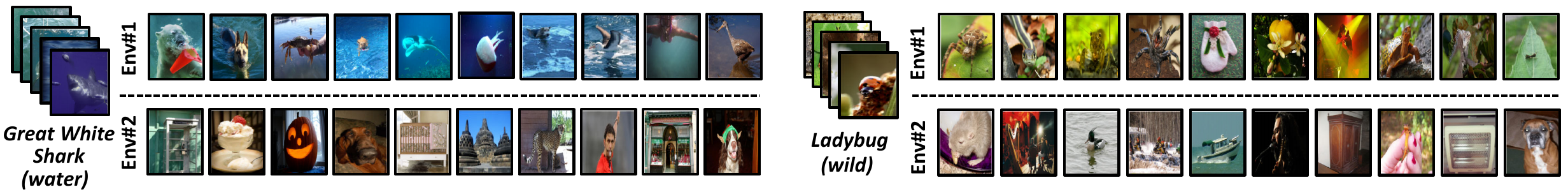}
    \vspace{-3mm}
    \caption{Visualizations of the top-10 images of generated environments for two classes (\ie, \texttt{great white shark} and \texttt{ladybug}) on VIPriors-10 dataset. We manually label their main context label (\ie, \texttt{water} and \texttt{wild}).}
    \label{fig:vis_cluster}
    \vspace{-1mm}
\end{figure}

\noindent\textbf{A4:} Recall that we have displayed the cluster results on a toy Colored MNIST data in Fig.~\ref{fig:cmnist_cluster} and validated the superiority of our similarity adjustment. Here we wonder how does it perform on real-world data with much comprehensive semantics? We visualize the top-10 images of Env\#1 and Env\#2 for two random selected classes in Fig.~\ref{fig:vis_cluster}. Interestingly, we can find that images of Env\#1 mainly share the context (\eg, \texttt{water}) with the \texttt{anchor} class (\eg, \texttt{Great White Shark}). In contrast, images of Env\#2 have totally different context. More importantly, the classes distribute almost uniformly in both Env\#1 and \#2, indicating that our adjusted similarity isolate the effect of the class feature.

\section{Conclusion}
We pointed out the theoretical reasons why learning from insufficient data is inherently more challenging than sufficient data---the latter will be inevitably biased to the limited environmental diversity. To counter such ``bad'' bias, we proposed to use two ``good'' inductive biases: equivariance and invariance, which are implemented as the proposed $\textsc{EqInv}$ algorithm. In particular, we used SSL to achieve the equivariant feature learning that wins back the class feature lost by the ``bad'' bias, and then proposed a class-wise IRM to remove the ``bad'' bias. 
For future work, we plan to further narrow down the performance gap by improving the class-muted clustering to construct more unique environments.

\noindent\textbf{Acknowledgement.} The authors would like to thank all reviewers for their constructive suggestions. This research is partly supported by AISG, A*STAR under its AME YIRG Grant (Project No.A20E6c0101).

%
%
\bibliographystyle{splncs04}
\bibliography{egbib,egbib_nips2021,egbib_iccv2021}

\newpage











\appendix
\noindent\textbf{\Large Appendix}
\vspace{0.2in}

This appendix is organized as follows:

\begin{itemize}
    
    \item[$\circ$] Section~\ref{sec:derivation} presents mathematical definition for the equivariant representation (Section~1) and derivation of the interventional ERM (Eq.~(1)).
    
    \item[$\circ$] Section~\ref{sec:implem} provides the implementation details of the Fig.~2 and Eq.~(6). We also elaborate the details of the used Colored MNIST dataset (Fig.~5) and the NICO dataset (Section~5.1).
    
    \item[$\circ$] Section~\ref{sec:additional_results} shows the results with MAE pretrained feature (Section~4); the attention map visualizations (Section~5.4) and the algorithm complexities (Section~5.4).
\end{itemize}


\section{Mathematical Definition \& Derivation}
\label{sec:derivation}

\subsection{The Mathematical Definition of Equivariance}

Let $\mathcal{U}$ be a set of (unseen) semantics, \eg, attributes such as ``digit'' and ``color''. There is a set of \emph{independent and causal mechanisms}~\cite{parascandolo2018learning} $\varphi: \mathcal{U}\to \mathcal{I}$, generating images from semantics, \eg, writing a digit ``0'' when thinking of ``0''~\cite{scholkopf2012on}. A \textbf{visual representation} is the inference process $\phi:\mathcal{I}\to \mathcal{X}$ that maps image pixels to vector space features, \eg, a neural network. We define \textbf{semantic representation} as the functional composition $f:\mathcal{U}\to \mathcal{I}\to\mathcal{X}$. 
Let $\mathcal{G}$ be the group acting on $\mathcal{U}$, \ie, $g\cdot u\in \mathcal{U}\times \mathcal{U}$ transforms $u\in \mathcal{U}$, \eg, a ``turn green'' group element changing the semantic from ``red'' to ``green''. 

\noindent\textbf{Definition 1.} (Equivariant Representation) 
\textit{Suppose there is a direct product decomposition $\mathcal{G}={g}_1\times\ldots\times {g}_m$ and $\mathcal{U}=\mathcal{U}_1\times\ldots\times \mathcal{U}_m$, where ${g}_i$ acts on $\mathcal{U}_i$ respectively. A feature representation is equivariant if there exists a group $\mathcal{G}$ acting on $\mathcal{X}$ such that:}
\begin{equation}
    f(g\cdot u)=g \cdot f(u), ~~\forall g\in \mathcal{G}, \forall u\in \mathcal{U}
\end{equation}
\textit{\eg, the feature of the changed semantic: ``red'' to ``green'' in $\mathcal{U}$, is equivalent to directly change the color vector in $\mathcal{X}$ from ``red'' to ``green''.}

As stated in Section~3.2, we follow the definition and implementation in~\cite{wang2021self} to achieve the sample-equvariant by using contrastive loss. Specifically, by assuming $\mathbf{x}\in\mathcal{X}$ as the feature, we can write the contrastive loss briefly as $\ell = -\log \frac{\exp (\mathbf{x}^T_i\mathbf{x}_j)}{\sum\nolimits_{\mathbf{x}\in\mathcal{X}}\exp(\mathbf{x}^T_j\mathbf{x})}$.
Then, if we use all the samples in the denominator of the loss, we can approximate to $\mathcal{G}$-equivariant features given limited training samples. This is because the loss minimization guarantees $\forall (\mathbf{x}_i, \mathbf{x}_j)\in\mathcal{X}\times\mathcal{X}, i\neq j\to \mathbf{x}_i\neq\mathbf{x}_j$.
We provide the proof in the following.

Suppose that the training loss $\ell$ is minimized, yet $\exists \mathbf{x}_a = \mathbf{x}_b\in\mathcal{X}$ for $a\neq b$. Let $\mathbf{x}_i \in \mathcal{X}$ in the denominator, and we have $\mathbf{x}_j^T \mathbf{x}_i=\mathrm{cos}(\theta_{i,j}) \norm{\mathbf{x}_i} \norm{\mathbf{x}_j}$, where $\theta_{i,j}$ is the angle between the two vectors. When $\mathbf{x}_i = \mathbf{x}_j$, $\mathrm{cos}(\theta_{i,j})=1$. So keeping $\norm{\mathbf{x}_i} \norm{\mathbf{x}_j}$ constant (\ie, the same regularization penalty such as L2), $\mathbf{x}_j^T \mathbf{x}_i$ can be further reduced if $\mathbf{x}_i \neq \mathbf{x}_j$, which reduces the training loss. This contradicts with the earlier assumption. Hence by minimizing the training loss, we can achieve sample-equivariant, \ie, different samples have different features. Note that this does not necessarily mean group-equivariant. However, the variation of training samples is all we know about the group action of $\mathcal{G}$, and we establish that the action of $\mathcal{G}$ is transitive on $\mathcal{X}$, hence we use the sample-equivariant features as the approximation of $\mathcal{G}$-equivariant features.

\subsection{Derivation of Eq.~(1)}

In this section, we will show the derivation for the backdoor adjustment formula using the three rules of \emph{do}-calculus~\cite{pearl2009causality}, whose detailed proof can be found in~\cite{pearl2009causality,pearl1995causal}.
For a causal directed acyclic graph $\mathcal{G}$, let $X, Y, Z$ and $W$ be arbitrary disjoint sets of nodes. We use $\mathcal{G}_{\overline X}$ to denote the manipulated graph where all incoming arrows to node $X$ are deleted. 
Similarly $\mathcal{G}_{\underline X}$ represents the graph where outgoing arrows from node $X$ are deleted. We use lower case $x,y,z$ and $w$ for specific values taken by each set of nodes: $X=x, Y=y, Z=z$ and $W=w$. For any interventional distribution compatible with $\mathcal{G}$, we have the following three rules:

\noindent\textbf{Rule 1} Insertion/deletion of observations.
If $(Y \independent Z | X,W)_{\mathcal{G}_{\overline X}}$:
\begin{equation}
    P(y|do(x),z,w)=P(y|do(x),w), 
\end{equation}

\noindent\textbf{Rule 2} Action/observation exchange.
If $(Y \independent Z | X,W)_{\mathcal{G}_{\overline X \underline Z}}$,
\begin{equation}
    P(y|do(x),do(z),w)=P(y|do(x),z, w), 
\end{equation}

\noindent\textbf{Rule 3} Insertion/deletion of actions.
If $(Y \independent Z | X,W)_{\mathcal{G}_{\overline X \overline {Z(W)}}}$,
\begin{equation}
    P(y|do(x),do(z),w)=P(y|do(x),w), 
\end{equation}
where $Z(W)$ is the set of nodes in $Z$ that are not ancestors of any $W$-node in $\mathcal{G}_{\overline{X}}$.

In our causal graph, the desired interventional distribution $P(Y|do(X))$ can be derived by:
\begin{align}
    P(Y|do(X)) 
    & = \sum_{z} P(Y|do(X),Z=z) P(Z=z|do(X)s) \label{bd1}\\
                        &= \sum_{z} P(Y|do(X),Z=z) P(Z=z) \label{bd2}\\
                        &= \sum_{z} P(Y|X,Z=z) P(Z=z) \label{bd3},
\end{align}
where Eq.~\eqref{bd1} follows the law of total probability; Eq.~\eqref{bd2} uses Rule 3 given $S \independent X$ in $\mathcal{G}_{\overline{X}}$;
Eq.~\eqref{bd3} uses Rule 2 to change the intervention term to observation as $(Y \independent X | Z)$ in $\mathcal{G}_{\underline{X}}$.
Therefore, by imposing Eq.~\eqref{bd3} into Eq.~(1) of the main paper, we can have: 
\begin{equation}
    \mathcal{R} = \sum_{x}\sum_{y}\sum_{z} \mathcal{L}(f(\phi(x)), y) P(y|x,z) P(z) P(x).
    \label{eq:intervention_appendix}
\end{equation}
\section{Implementation Details}
\label{sec:implem}

\begin{figure}[t]
\captionsetup{font=footnotesize,labelfont=footnotesize}
    \centering
    \includegraphics[width=1.0\textwidth]{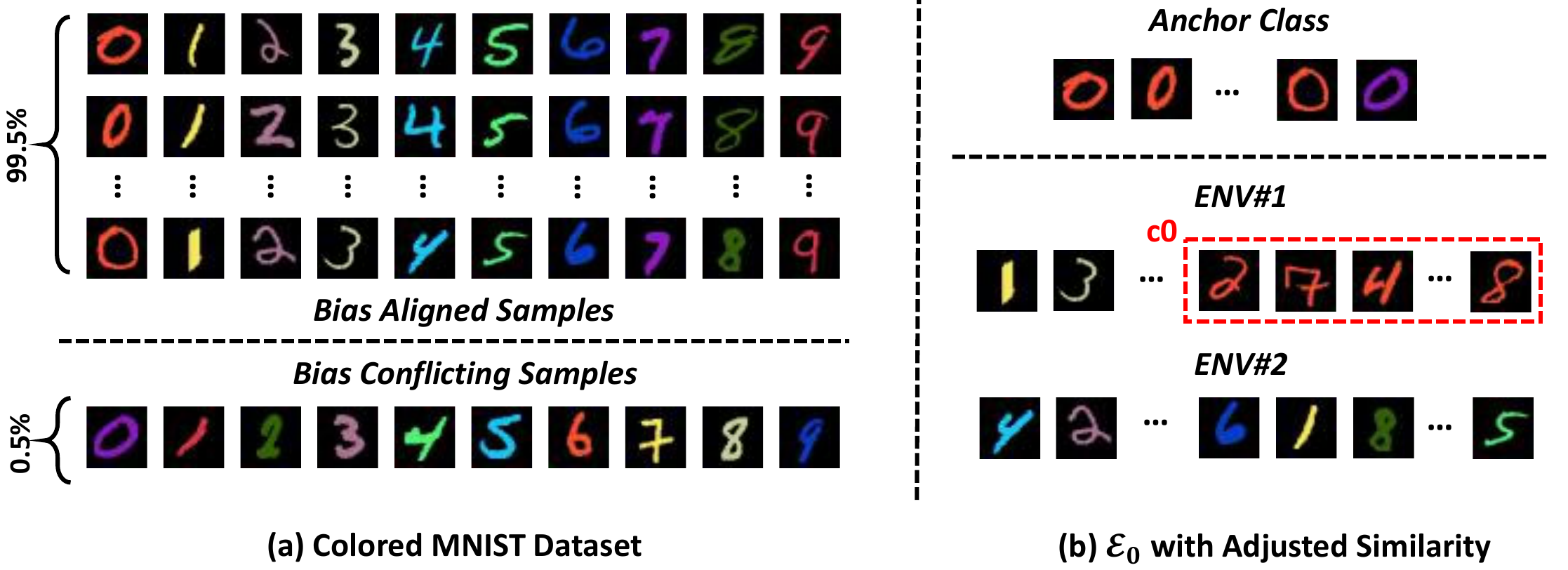}
    \caption{We illustrate (a) the example images of the Colored MNIST dataset; (b) the generated environment $\mathcal{E}_0$ with adjusted similarity.}
    \vspace{-3mm}
    \label{fig:cmnist_data}
\end{figure}

\subsection{Implementation Details of Fig.~2}

\noindent\textbf{The t-SNE Visualization.}
We adopt the t-SNE visualization here to reflect the true data distribution and expect the feature extraction model to be able to accurately structure the relationships between images. 
Recently, Contrastive Language-Image Pretraining (CLIP)~\cite{radford2021learning} is proposed for solving vision tasks by exploiting contrastive learning with very large-scale noisy image-text pairs. Such large data makes it nearly a ``Sufficient Data Situation''. And the conventional ERM algorithm can achieve optimal performance, as we introduced in Section~3 of the main paper. Indeed, CLIP achieves inspirational performances on various visual classification tasks. In this paper, we utilize the CLIP pretrained backbone (``ViT-Base/32'') to extract feature of the training and testing images of class \texttt{Hen} which is randomly chosen. Then we draw t-SNE visualization with the open-source codebase\footnote{https://github.com/DmitryUlyanov/Multicore-TSNE}.

\noindent\textbf{The Test Accuracy.}
We trained a ResNet-50 model from scratch on each training set and evaluated on testing images. We calculated accuracy on class \texttt{Hen}.

\subsection{Implementation Details of Eq.~(6)}

In Section~4 Step 3, we propose the class-wise IRM to regularize the invariance within each class as stated in Eq.~(6). In practice, we adopt a more practical version named REx~\cite{krueger2020out} of Eq.~(6) which improves the vanilla IRM in the covariate shift situation.
Specifically, \cite{krueger2020out} discards the dummy classifier $w$ and changes the penalty term of IRM to the variance of risks as the regularization:
\begin{equation}
    \mathcal{L}_k = \lambda \mathrm{Var(\{\ell(1),...,\ell(e)\})} + \sum\nolimits_{e\in\mathcal{E}_k} \ell (e).
    \label{eq:supcont_irm_appendix}
\end{equation}
Same as IRM, REx aims to encourage the invariance across different environments, but provides a simpler, stabler and more effective implementation~\cite{krueger2020out}.

\subsection{Details of Colored MNIST dataset in Fig.~5}

Figure~\ref{fig:cmnist_data} (a) shows the example images of the Colored MNIST~\cite{nam2020learning} dataset. As we introduced in Section~4 of the main paper, the Colored MNIST dataset injects \textit{color} bias on each digit. There are 99.5\% bias-aligned samples and only 0.5\% images are non-bias samples. For example, most of \texttt{0} are red. Figure~\ref{fig:cmnist_data} (b) illustrates the generated environment $\mathcal{E}_0$ of \texttt{anchor} class \texttt{0} with the adjusted similarity using real images. We can clearly observe that the biased color \texttt{c0} (\ie, red) of digit \texttt{0} distributes differently in Env\#1 and Env\#2, while other semantics keep invariant. This will encourages the bias color to be removed during the following class-wise IRM process.

\begin{table}[t]
\captionsetup{font=footnotesize,labelfont=footnotesize,skip=1pt}
\centering
\caption{Recognition accuracies (\%) on the VIPriors-10 and NICO datasets with ViT-Base/16 as the feature backbone and MAE~\cite{he2021masked} as the SSL method. ``Aug.'' represents augmentation. Our results are highlighted.}
\scalebox{0.85}{
\begin{tabular}{p{4.0cm}p{1.1cm}<{\centering}p{1.1cm}<{\centering}p{1.1cm}<{\centering}p{1.1cm}<{\centering}}
\toprule\toprule
\multicolumn{1}{c}{\multirow{2}{*}{Model}} & \multicolumn{2}{c}{VIPriors-10} & \multicolumn{2}{c}{NICO} \\ \cmidrule(lr){2-3} \cmidrule(lr){4-5}
\multicolumn{1}{c}{}  & ~Val~ & ~Test~ & ~Val~ & ~Test~ \\ \midrule
\textit{Train from Scratch} &  &  &  &  \\
Baseline (ViT-Base/16) & 4.74 & 4.50 & 32.23 & 31.46 \\
Random Aug.~\cite{cubuk2020randaugment} & 5.40 & 4.92 & 33.54 & 31.92 \\ \midrule
\textit{Train from SSL} &  &  &  &  \\
MAE~\cite{he2021masked} & 16.04 & 14.63 & 54.10 & \textbf{52.29} \\
~~+ \algname (Ours) & \cellcolor{mygray}{\textbf{16.93}} & \cellcolor{mygray}{\textbf{15.48}} & \cellcolor{mygray}{\textbf{56.26}} & \cellcolor{mygray}{\textbf{52.29}} \\ \midrule
MAE~\cite{he2021masked} + Random Aug.~~ & 16.70 & 15.39 & 56.11 & 52.91 \\
~~+ \algname (Ours) & \cellcolor{mygray}{\textbf{17.53}} & \cellcolor{mygray}{\textbf{16.00}} & \cellcolor{mygray}{\textbf{57.96}} & \cellcolor{mygray}{\textbf{54.14}} \\ 
\bottomrule\bottomrule
\end{tabular}}
\label{tab:mae}
\end{table}

\subsection{Experimental Details}

\noindent\textbf{More Implementation Details of the Equivariant Learning.}
As stated in the main paper, we utilize different Self-Supervised Learning techniques in Step 1 Equivariant Learning. For implementation, we train for 800 epochs using ResNet-50/-18 and ViT-Base/16 as the encoder architecture. We just follow the original methods to use the default parameters and training schedules except for some slight changes to adapt to the VIPriors and NICO dataset. Specifically, we set the queue size as 16384 for MoCo-v2~\cite{he2019moco,chen2020mocov2}. For the NICO dataset, we train the MAE for 2000 epochs and adopt the mixup version of IP-IRM~\cite{wang2021self} to achieve the reasonable performance. Moreover, please note that we follow the other team's solution~\cite{zhao2020distilling} of VIPriors Challenge to use both train and val set for the Stage 1 SSL pretraining with no need of the label for all the comparison methods. Then for the second fine-tuning stage, as stated in the main paper, we only use the train set images and labels for training.

\noindent\textbf{More Implementation Details of the Downstream Fine-tuning.}
In the main paper, we have introduced most training schedules for ResNet model.
Besides the training schedules introduced in the main paper, we set $\lambda=2, 10, 100$ for VIPriors-10, 20, 50 dataset. This parameter choice also makes sense from intuitive since the demand of the invariance regularization is decreasing with more training samples. 
Please note that, for fair comparison with data-efficient learning methods, we did not apply any strong data augmentation in our downstream training (after SSL), even though it is common in SSL works.
For MAE with ViT-Base/16, we follow the default end-to-end fine-tuning schedule: AdamW as the optimizer with base learning rate 5e-4 using the cosine learning rate decay; the layer-wise learning rate decay is set to 0.65 and weight decay is set to 0.05; the drop path is set to 0.1 and the warmup epochs are 5. We decrease the batch size to 256 and not use the advanced augmentation (\ie, cutmix, mixup, label smoothing) to keep consistent with the ResNet model.
For our proposed class-wise IRM, the hidden size of the MLP $g$ is 512 with batch normalize layer and ReLU activation. The output dimension of $g$ is 128, same with SimCLR~\cite{chen2020simple}. We also utilize the weight normailzation~\cite{salimans2016weight} on the fc layer $f$ for the stable training.
\section{Additional Results}
\label{sec:additional_results}

\subsection{Results with MAE~\cite{he2021masked} feature}

\begin{figure}[t]
\captionsetup{font=footnotesize,labelfont=footnotesize}
    \centering
    \includegraphics[width=1.0\textwidth]{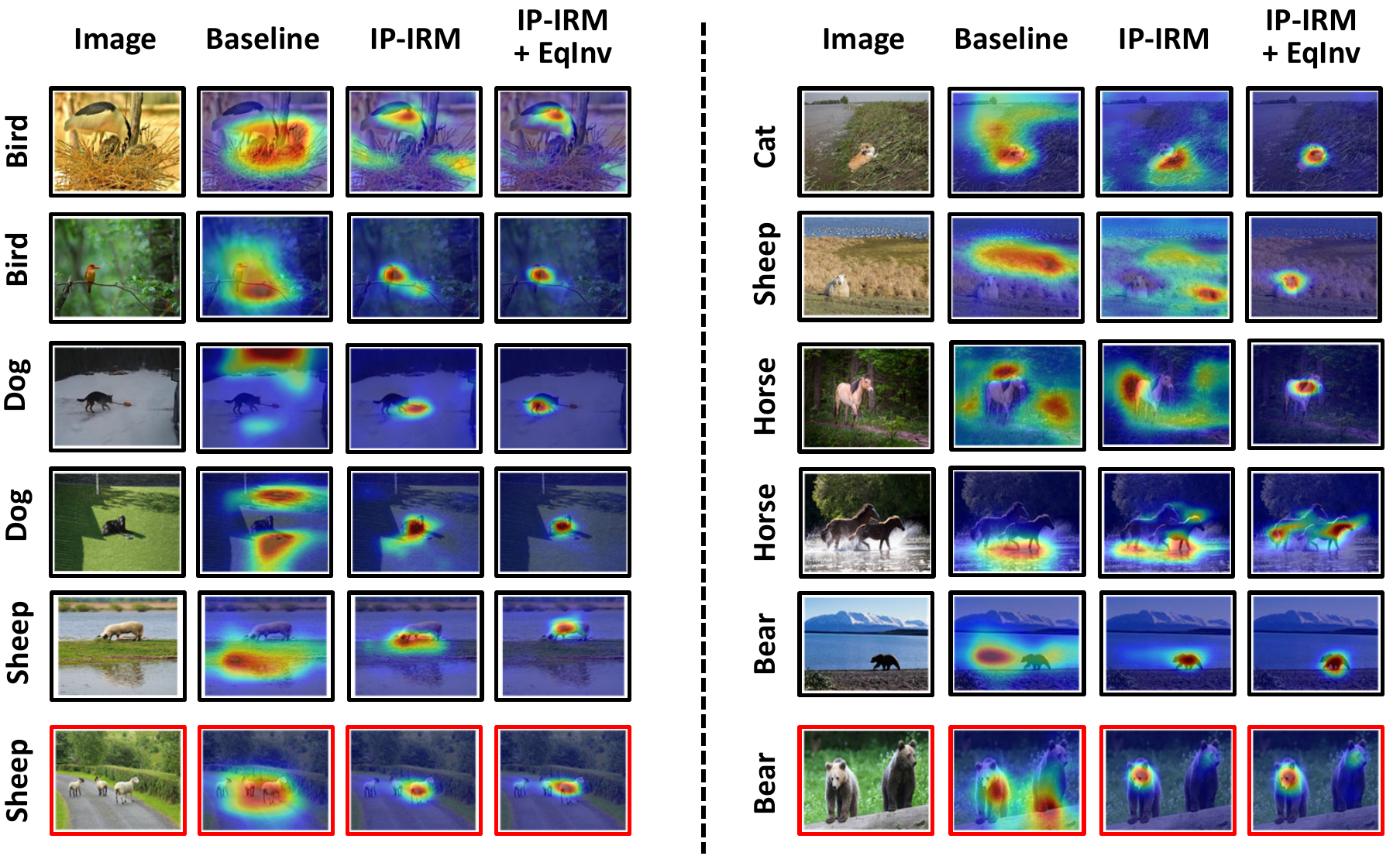}
    \caption{Visual attention visualizations on NICO dataset with our proposed \algname and baseline methods. We adopt IP-IRM~\cite{wang2021self} as the SSL method. The red box represents the failure case.}
    \label{fig:attn_nico}
\end{figure}

Table~\ref{tab:mae} shows the recognition accuracies on the VIPriors-10 and NICO datasets with ViT-Base/16 as the feature backbone and MAE~\cite{he2021masked} as the SSL method. Similar to the Table~1 in the main paper, we can observe that compared to training from scratch, both imposing equivariance and invariance inductive bias can boost the performance. 
However, we also find that the improvements of considering invariance inductive bias are not such huge compared to that of the ResNet structure. The possible reason is the Visual Transformer~\cite{dosovitskiy2020image,khan2021transformers} structure itself is more robust~\cite{bhojanapalli2021understanding,bai2021transformers} to the distribution shift than the CNN model (\eg, ResNet). That is, the self-attention-like architectures of Visual Transformer have partially achieved the invariance.

\subsection{Algorithm Complexities}
\begin{table}[h]
\captionsetup{font=footnotesize,labelfont=footnotesize,skip=1pt}
\centering
\caption{The model size and computational cost comparison between our proposed \algname and baseline models with different feature backbones.}
\scalebox{0.8}{
\begin{tabular}{lcccc}
\toprule\toprule
Model & ~Params (M)~ & ~Flops (G)~ & ~MACs (G)~ & ~Time (s)~ \\ \midrule
ResNet-18 & 11.180 & 3.644 & 1.822 & 0.025 \\
~~+ \algname (Ours) & 11.510 & 3.646 & 1.823 & 0.105 \\
ResNet-50 & 25.560 & 8.244 & 4.122 & 0.061 \\
~~+ \algname (Ours) & 26.680 & 8.246 & 4.123 & 0.342 \\
ViT-Base/16 & 86.570 & 33.712 & 16.856 & 0.065 \\
~~+ \algname (Ours) & 87.030 & 33.712 & 16.856 & 0.297 \\ \bottomrule\bottomrule
\end{tabular}}
\label{tab:complex}
\end{table}

We show the model sizes and the computational costs in Table~\ref{tab:complex}. The ``Time (s)'' denotes the forwarding process time with $bs$ images as input. $bs$ is set to $128$ for ResNet and $64$ for ViT, based on GPU memory consumption. 
We can see that compared to baseline models, deploying \algname does not cause many extra network parameters and computation costs. This is because in our \algnamens, the invariance inductive bias is implemented by only one learnable mask layer and one MLP layer, bringing little overhead.

\subsection{Attention Visualizations}

\begin{figure}[t]
\captionsetup{font=footnotesize,labelfont=footnotesize}
    \centering
    \includegraphics[width=0.6\textwidth]{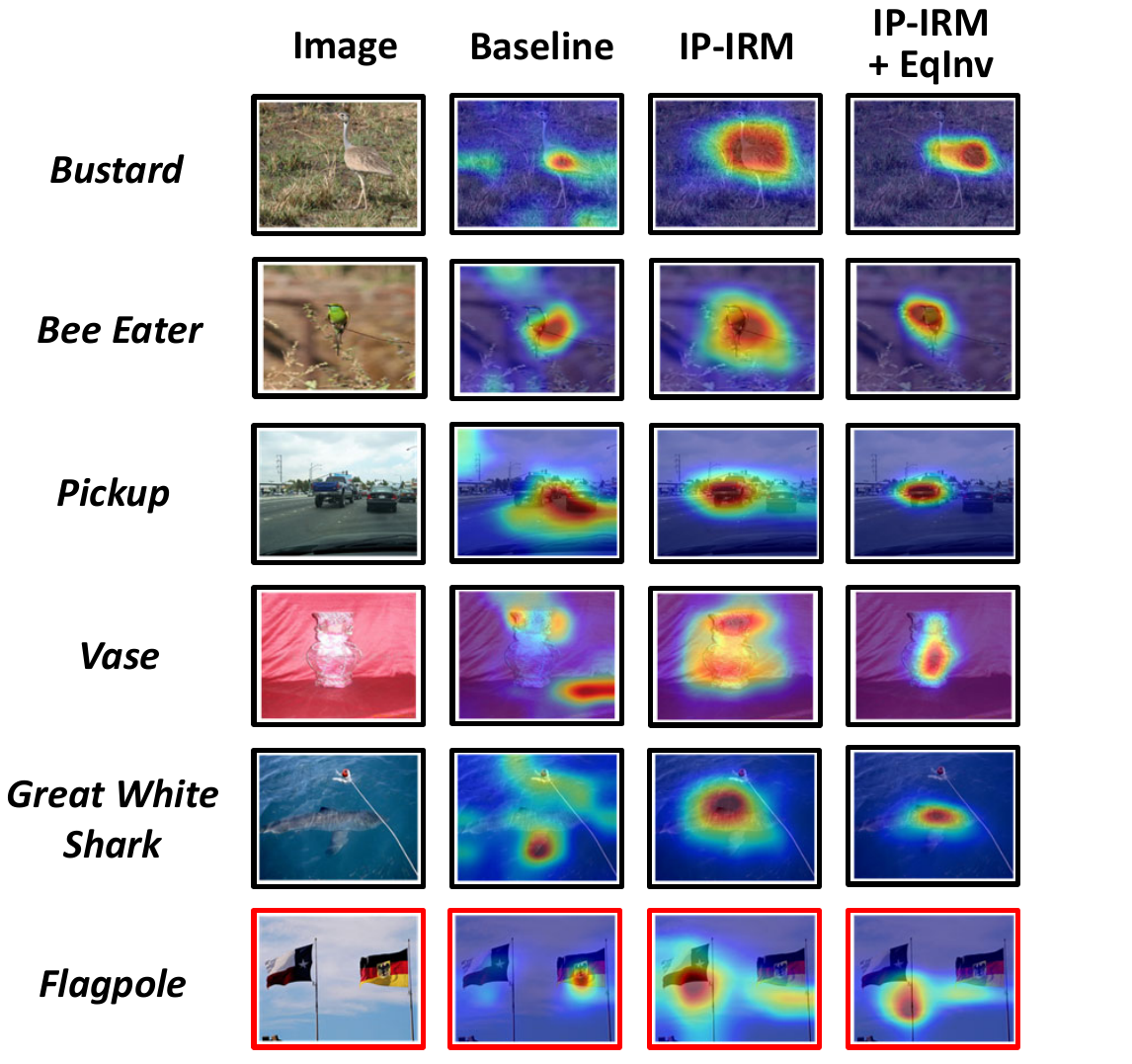}
    \caption{Visual attention visualizations on VIPriors-10 dataset with our proposed \algname and baseline methods. We adopt IP-IRM~\cite{wang2021self} as the SSL method. The red box represents the failure case.}
    \label{fig:attn_vip10}
\end{figure}

Figure~\ref{fig:attn_nico} and~\ref{fig:attn_vip10} show the qualitative attention map comparisons between baseline (\ie, training from scratch), incorporating SSL pretraining and our proposed \algnamens. We utilize CAM~\cite{zhou2016learning} for the visualization.
We can clearly observe that:
\begin{itemize}
    \item[$\circ$] Training from scratch (the second column) produces many inaccurate attentions, even totally misses the object area (\eg, the last three rows of Fig.~\ref{fig:attn_nico} left). This indicates the severe environmental bias of the model trained with the insufficient samples (\textit{cf.} Section~3.1 of the main paper).
    
    \item[$\circ$] Though incorporating SSL pretraining (the third column) greatly alleviates such problem by imposing the equivariance inductive bias, the model still attends to partial context area. It means the model may still be confounded by the environmental feature during the downstream fine-tuning.
    
    \item[$\circ$] By further imposing the invariance inductive bias with our proposed \algname (the last column), the model can achieve much more accurate and tighter attention focusing on the object area. We also display the failure cases in red boxes. We can find that our \algname cannot accurately attend to multiple objects (\eg, three sheeps and two bears in Fig.~\ref{fig:attn_nico}) or small objects (\eg, the flagpole in Fig.~\ref{fig:attn_vip10}). But our \algname can still achieve relatively better attention maps compared to comparison methods. We will explore such failure cases in the future work.
\end{itemize}


\end{document}